\documentclass[letterpaper]{article} 
\usepackage{aaai23}  
\usepackage{times}  
\usepackage{helvet}  
\usepackage{courier}  
\usepackage[hyphens]{url}  
\usepackage{graphicx} 
\usepackage{natbib}  
\usepackage{caption} 
\usepackage{algorithm}
\usepackage{algorithmic}

\usepackage{newfloat}
\usepackage{listings}

\usepackage{amsmath,amsfonts,bm,amssymb,multirow}

\DeclareCaptionStyle{ruled}{labelfont=normalfont,labelsep=colon,strut=off} 
\floatstyle{ruled}
\newfloat{listing}{tb}{lst}{}
\floatname{listing}{Listing}
\title{Learning by Applying: A General Framework for Mathematical Reasoning\\ via Enhancing Explicit Knowledge Learning}
\author{
    Jiayu Liu\textsuperscript{\rm 1, \rm 2},
    Zhenya Huang\textsuperscript{\rm 1, \rm 2}\thanks{Corresponding Author.},
    Chengxiang Zhai\textsuperscript{\rm 3},
    Qi Liu\textsuperscript{\rm 1, \rm 2}
}
\affiliations{
    \textsuperscript{\rm 1}Anhui Province Key Laboratory of Big Data Analysis and Application, School of Data Science \& \\School of Computer Science and Technology, University of Science and Technology of China\\
    \textsuperscript{\rm 2}State Key Laboratory of Cognitive Intelligence\\
    \textsuperscript{\rm 3}University of Illinois at Urbana-Champaign\\
    jy251198@mail.ustc.edu.cn,\ \{huangzhy,\ qiliuql\}@ustc.edu.cn,\ czhai@illinois.edu
}

\usepackage{bibentry}
\begin{document}

\maketitle

\begin{abstract}
Mathematical reasoning is one of the crucial abilities of general artificial intelligence, which requires machines to master mathematical logic and knowledge from solving problems. However, existing approaches are not transparent (thus not interpretable) in terms of what knowledge has been learned and applied in the reasoning process. In this paper, we propose a general Learning by Applying (LeAp) framework to enhance existing models (backbones) in a principled way by explicit knowledge learning. In LeAp, we perform knowledge learning in a novel \emph{problem-knowledge-expression} paradigm, with a Knowledge Encoder to acquire \emph{knowledge} from \emph{problem} data and a Knowledge Decoder to apply \emph{knowledge} for \emph{expression} reasoning. The learned mathematical knowledge, including word-word relations and word-operator relations, forms an explicit knowledge graph, which bridges the knowledge ``learning'' and ``applying'' organically. Moreover, for problem solving, we design a semantics-enhanced module and a reasoning-enhanced module that apply knowledge to improve the problem comprehension and symbol reasoning abilities of any backbone, respectively. We theoretically prove the superiority of LeAp's autonomous learning mechanism. Experiments on three real-world datasets show that LeAp improves all backbones' performances, learns accurate knowledge, and achieves a more interpretable reasoning process.
\end{abstract}

\section{Introduction}
\begin{figure}[t]
\centering
\includegraphics[width=\linewidth]{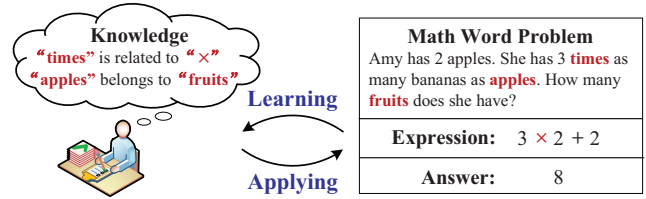}
\caption{Learning and applying knowledge for MWP.}
\label{intro}
\end{figure}

Mathematical reasoning is one of the core abilities and signs of the intelligence level of general artificial intelligence~\cite{zhang2020gap}. It requires machines to grasp mathematical knowledge and logical thinking from solving several mathematical problems~\cite{lewkowycz2022solving,seo2015solving}. Among them, we specifically study math word problems (MWP) in this paper, which is a fundamental reasoning task that has attracted much attention since the 1960s~\cite{feigenbaum1963computers}. Figure~\ref{intro} illustrates a toy example. Generally, a math word problem is represented as a problem sentence (``Amy has ... have?'') that poses a question requesting an unknown quantity. To solve it, the machine needs to understand the verbal description that contains words (e.g.,``Amy'') and quantities (e.g.,``2''), and then reason a mathematical expression (i.e.,``$3\times 2+2$''), based on which finally get the answer (i.e.,``8'').

In the literature, traditional MWP solvers include rule-based, statistic-based, and semantics parsing-based~\cite{zhang2020gap}. In recent years, sequence to sequence (Seq2Seq) framework has thrived in MWP task~\cite{wang2017deep,lan2022mwptoolkit}, following a \emph{problem-expression} paradigm to translate problems into expressions, where the existing work has focused on improving MWP in two ways, including promoting problem understanding~\cite{lin2021hms,kim2022improving} and improving expression reasoning~\cite{zhang2020teacher,cao2021bottom}. However, such a \emph{problem-expression} paradigm is still far from encompassing human-like mathematical reasoning capacity, because it lacks the processes of learning and applying explicit knowledge. On one hand, according to the educational theory of cognitivism~\cite{mowrer1960learning,muhajirah2020basic}, humans acquire explicit mathematical knowledge from solving problems~\cite{liu2019ekt}, e.g., the word ``times'' is related to the operator ``$\times$'' and ``apples'' belongs to ``fruits'' from the problem in Figure~\ref{intro}. On the other hand, humans produce solutions for MWP by applying this knowledge in logical thinking~\cite{liang2018symbolic}. It is necessary to integrate these two processes organically into machines to build a stronger AI~\cite{tsatsou2021towards,de2011neural}.

From this perspective, existing MWP solvers have some room for improvement. First, they ignore the learning process of explicit knowledge. To be specific, the learned knowledge of existing solvers is implicitly contained in the parameters and network architectures, which is not transparent to humans. Second, the lack of applying explicit knowledge (e.g., ``times'' is related to ``$\times$'' in Figure~\ref{intro}) hinders their reasoning ability and interpretability of how they reason the answers. More importantly, we emphasize that both knowledge learning and applying are general human capabilities~\cite{huang2020learning} that can benefit different MWP solvers. To this end, this paper aims to construct a general framework in which different MWP solvers can learn and apply explicit knowledge to achieve better reasoning ability.

However, many challenges remain to be confronted. First, there is no clear scheme of how to formalize the explicit knowledge that is helpful for MWP and applicable to different solvers. Second, it is challenging to probe a learning mechanism that simulates how humans gain knowledge from solving MWP, which meanwhile should be general to work with different solvers. Third, we need to design a general knowledge application mechanism based on distinct solver architectures, which is underexplored nowadays.

To address these challenges, we propose a novel framework named Learning by Applying (LeAp) for MWP, which adopts a \emph{problem-knowledge-expression} architecture. In LeAp, we define two types of explicit and general mathematical knowledge, including word-word relations and word-operator relations. Then, we implement LeAp by a Variational AutoEncoder (VAE) that contains a Knowledge Encoder and a Knowledge Decoder. Specifically, Knowledge Encoder acquires \emph{knowledge} from \emph{problem} sentences, and Knowledge Decoder applies the learned \emph{knowledge} to reason corresponding \emph{expressions}. The combination of these two components constitutes our novel ``learning knowledge by applying it'' mechanism. The learned knowledge explicitly forms a knowledge graph in the middle and serves as a bridge connecting the two components, which is transparent. Moreover, for MWP solving, we propose a semantics-enhanced module and a reasoning-enhanced module in Knowledge Decoder, which apply knowledge to promote problem comprehension and symbol reasoning, respectively. Our LeAp is a general framework that benefits existing MWP solvers by improving their reasoning abilities. We conduct extensive experiments by instantiating several backbone solvers in LeAp. The experimental results on three datasets show the improvements of LeAp on answer reasoning, effect of knowledge learning, as well as reasoning interpretability. The contributions of this paper are as follows:
\begin{itemize}
  \item We propose a general Learning by Applying (LeAp) framework to learn and apply explicit knowledge, where existing MWP solvers can serve as its backbone and benefit from it to improve reasoning ability.
  \item We design a novel semantics-enhanced module and a novel reasoning-enhanced module for knowledge application, which enhance existing MWP solvers on both answer accuracy and reasoning interpretability.
  \item We theoretically analyze the superiority of our autonomous knowledge learning mechanism in LeAp and experimentally validate its effectiveness.
\end{itemize}

\section{Related Work}\label{related_work}
In this section, we summarize the related work as follows.

\textbf{Math Word Problems.} Early efforts to solve MWP range from rule-based methods~\cite{bakman2007robust,fletcher1985understanding}, statistic-based methods~\cite{hosseini2014learning,mitra2016learning}, to semantics parsing-based methods~\cite{koncel2015parsing,shi2015automatically}. They are characterized by relying on manually crafted rules, machine learning models, and semantic structure of problems, respectively. Recently, Wang et al.~\shortcite{wang2017deep} first proposed a seq2seq model that translated the problem sentence into expression, following a \emph{problem-expression} paradigm. Based on such a manner, we summarize advanced methods into two categories: semantics-focused and reasoning-focused. Specifically, semantics-focused methods aim to promote the comprehension of problems with advanced networks (e.g., graph neural networks~\cite{zhang2020graph}, pre-trained language models~\cite{liang2021mwp,shen2021generate,kim2022improving,yu2021improving,huang2021disenqnet}), or additional information~\cite{lin2021hms,wu2021edge,huang2020neural}. For example, Zhang et al.~\shortcite{zhang2020graph} proposed Graph2Tree to capture the relationships and order information among quantities. Reasoning-focused methods aim to improve the expression reasoning process~\cite{wang2019template,shen2020solving,cao2021bottom,jie2022learning}, such as GTS~\cite{xie2019goal} that applied the goal-driven decomposition mechanism to reason an expression tree. Besides, Shen et al.~\shortcite{shen2020solving} produced an ensemble of multiple encoders and decoders, combining their advantages in both semantic understanding and reasoning.

\textbf{Knowledge Learning and Applying.} One of our most crucial targets is that LeAp can learn and apply explicit knowledge for MWP. Thus, we summarize the related work regarding knowledge learning and knowledge applying, respectively. For knowledge learning, it expects machines to gain knowledge from data~\cite{de2011neural,labhishetty2022differential}. Since our knowledge in LeAp forms an explicit knowledge graph, a relevant task is link prediction~\cite{chen2021topology,cai2020multi} (or knowledge graph completion~\cite{bansal2019a2n,cheng2021uniker}), which generally learns unknown knowledge (edges) in a knowledge graph with the existing edges. For example, Bordes et al.~\shortcite{bordes2013translating} considered the semantics of knowledge and interpreted it as translation operation. Pei et al.~\shortcite{pei2019geom} captured the structure information and the long-range dependencies via a novel geometric perspective. Some researchers also investigate learning other types of knowledge, e.g., background knowledge~\cite{peyrard2020klearn}, logic knowledge~\cite{dai2021abductive}, and implicit knowledge in pre-trained language models~\cite{petroni2019language}. For knowledge applying, various types of knowledge have been applied in many machine learning tasks, such as conversation generation~\cite{zhou2018commonsense}, question answering~\cite{liu2021neural}, and recommender systems~\cite{xia2021knowledge}. Some special forms of knowledge (e.g., logic rules~\cite{qu2019probabilistic}, mathematical property~\cite{pei2020curvature}) have also played an important role in many studies. We refer the readers to a more detailed survey conducted by Laura von Rueden et al.~\shortcite{von2021informed}. Especially, Wu et al.~\shortcite{wu2020knowledge} made a basic attempt to incorporate an external manually constructed knowledge base for MWP task.
\begin{figure*}[t]
	\centering
	\includegraphics[scale=0.74]{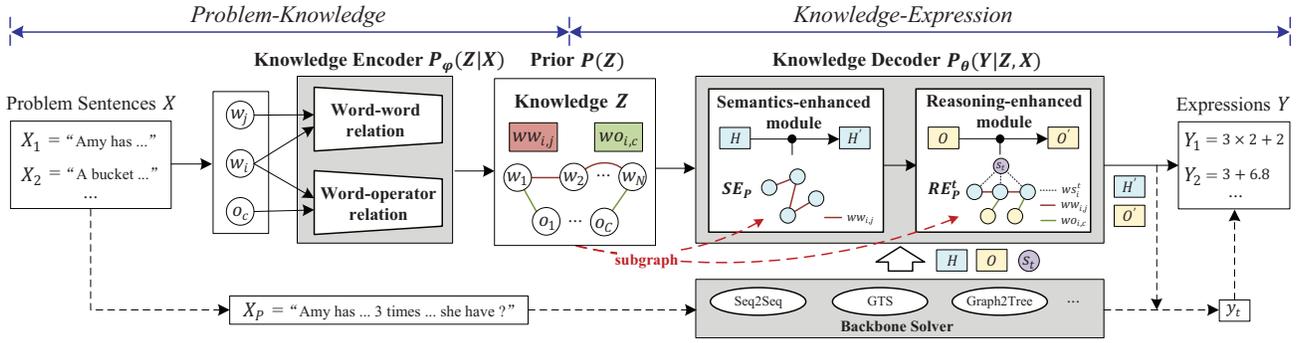}
	\caption{The architecture of LeAp, which operates in a \emph{problem-knowledge-expression} paradigm.}
	\label{leap}
\end{figure*}

Different from previous studies on knowledge learning, our LeAp gains mathematical knowledge by applying it to reason answers. Compared with existing work on MWP, it is a general framework that empowers different solvers with explicit knowledge learning and applying in a novel \emph{problem-knowledge-expression} paradigm, further improving their reasoning abilities. Besides, we propose a semantics-enhanced module and a reasoning-enhanced module in LeAp to apply knowledge for MWP solving, which bring better answer accuracy and reasoning interpretability.
\section{LeAp: Learning by Applying}
In this section, we first formally define our problem and then introduce the architecture of our LeAp in details.
\subsection{Problem Definition}
A MWP dataset is denoted as $D=(X, Y)$, where $X$ is the set of problem sentences and $Y$ is the set of corresponding expressions. Specifically, $X_P=\{w_1,...,w_n\}\in X$ is a sequence of $n$ word tokens and numeric values of problem $P$, where $w_i$ is either a word token (e.g., ``Amy'' in Figure~\ref{intro}) or a numeric value (e.g., ``2''). $Y_P=\{y_1,...,y_m\}\in Y$ is a sequence of $m$ symbols. Each symbol $y_i$ comes from a target vocabulary $V_P$ composed of the operator set $V_O$ (e.g., $\{+,\times,-,\div\}$), numeric constant set $V_N$ (e.g., $\pi$), and numeric values $N_P$ in $X_P$, i.e., $V_P=V_O\cup V_N\cup N_P$. Note that different problems may have different $V_P$ since $N_P$ varies with $P$. The goal of MWP is to train a solver that reads the problem sentence $X_P$, generates a valid mathematical expression $Y_P$, and gets a numeric answer $a_P$ based on $Y_P$.

The knowledge $Z$ we consider is explicitly represented as a mathematical knowledge graph. Its vertices include words and operators, and the knowledge we focus on is the existence of its edges. Specifically, we formalize $Z$ as word-word relation $ww_{i,j}$ and word-operator relation $wo_{i,c}$, i.e., $Z=\{ww_{i,j}, wo_{i,c}|i,j=1,...,N;c=1,...,C\}$, where $N$ and $C$ are the number of words and operators in MWP task. $ww_{i,j}$ describes the relationship between words $w_i$ and $w_j$ (e.g., ``apples'' belongs to ``fruits''), while $wo_{i,c}$ captures the one between word $w_i$ and operator $o_c$ (e.g., ``times'' is related to ``$\times$''). Both $ww_{i,j}$ and $wo_{i,c}$ are set as binary variables, which equal $1$ if there exists knowledge between word $w_i$ and word $w_j$ (or operator $o_c$), and $0$ otherwise. Please note that our formulation can be easily extended to incorporate multiple relationships (edges) between words and operators, such as hypernymy and antonymy~\cite{shehata2009wordnet}.

Our goal is to build a framework that (1) learns mathematical knowledge $Z$ from solving MWP; (2) applies knowledge $Z$ to reason the answers for MWP. These two goals are coupled with each other and achieved collaboratively.
\subsection{LeAp Architecture}
Intuitively, the educational theory of cognitivism~\cite{mowrer1960learning,muhajirah2020basic} indicates that a learner fosters a strong connection to the knowledge (e.g., ``times'' is related to ``$\times$'') by directly applying it (e.g., solve the problem in Figure~\ref{intro}). Drawing this insight, we construct our LeAp framework with a novel \emph{problem-knowledge-expression} architecture to achieve the mechanism of ``learning knowledge by applying it''. Specifically, the \emph{problem-knowledge} process acquires \emph{knowledge} from \emph{problem} data, and the \emph{knowledge-expression} process applies this \emph{knowledge} to reason \emph{expressions} for answers, which further guides to learn reasonable knowledge autonomously. To this end, we formalize LeAp as a Variational AutoEncoder (VAE)~\cite{kingma2013auto}, which is shown in Figure~\ref{leap}. It consists of three main parts: (1) a Knowledge Encoder $P_{\varphi}(Z|X)$ that acquires knowledge $Z$ from problem sentences $X=\{X_P\}$ (\emph{problem-knowledge}); (2) a Knowledge Decoder $P_{\theta}(Y|Z,X)$ that reasons the expressions $Y=\{Y_P\}$ based on $X$ and $Z$ (\emph{knowledge-expression}); (3) a knowledge prior $P(Z)$ of $Z$. The training objective of LeAp is to maximize the Evidence Lower Bound (ELBO):
\begin{equation}\label{elbo}
\small
L=\underbrace{\mathrm{E}_{P_{\varphi}(Z|X)}\left[\log P_{\theta}(Y|Z, X)\right]}_{L_1}-\underbrace{\mathrm{KL}\left(P_{\varphi}(Z|X) \| P(Z)\right)}_{L_2},
\end{equation}
where the first term $L_1$ optimizes the performance of MWP solving, and the second term $L_2$ regularizes the knowledge learning results. In the following, we will introduce the details of Knowledge Encoder $P_{\varphi}(Z|X)$, Knowledge Decoder $P_{\theta}(Y|Z,X)$, and the knowledge prior $P(Z)$ in turn.
\subsubsection{Knowledge Encoder.}
Knowledge Encoder $P_{\varphi}(Z|X)$ aims at acquiring explicit knowledge $Z$ from the problem sentence set $X$ as shown in Figure~\ref{leap}, which operates the \emph{problem-knowledge} process. Since $ww_{i,j}, wo_{i,c}\in Z$ are binary, we map each word $w_i$ and operator $o_c$ to the vector $\bm{w}_i,\bm{o}_c \in \mathbb{R}^d$ respectively ($d$ is the dimension), and feed $\bm{w}_i,\bm{w}_j,\bm{o}_c$ into different networks to encode their Bernoulli distributions. Formally, we model Knowledge Encoder as:
\begin{small}
\begin{equation}\label{know_enc}
\begin{aligned}
&P_{\varphi}(ww_{i,j}|X) = \operatorname{Bernoulli}\left(\sigma\left(f_{1}\left([\bm{w}_{i}, \bm{w}_{j}]\right)\right)\right), \\
&P_{\varphi}(wo_{i,c}|X) = \operatorname{Bernoulli}\left(\sigma\left(f_{2}\left([\bm{w}_{i}, \bm{o}_{c}]\right)\right)\right),
\end{aligned}
\end{equation}
\end{small}

\noindent where $\sigma$ is the sigmoid function, $f_{1}$ and $f_{2}$ are neural networks that transform the concatenation $[\cdot]$ of $\bm{w}_{i}, \bm{w}_{j}$ and $\bm{w}_{i}, \bm{o}_{c}$, respectively. Note that $f_{1},f_{2}$ can be implemented ranging from MLP to pre-train language models~\cite{petroni2019language}. Since we focus more on learning knowledge from MWP, we do not emphasize their difference and adopt MLP~\cite{kipf2018neural} for simplicity.

During training, Knowledge Encoder $P_{\varphi}(Z|X)$ needs to sample $Z$ to estimate $\mathrm{E}_{P_{\varphi}(Z|X)}\left[\log P_{\theta}(Y|Z, X)\right]$, i.e., $L_1$ in Eq.~(\ref{elbo}). However, it is hard to use reparameterization to backpropagate the derivatives since $ww_{i,j},wo_{i,c}\in Z$ are binary~\cite{kipf2018neural}. Therefore, we adopt a continuous approximation of Eq.~(\ref{know_enc}) when optimizing Eq.~(\ref{elbo}):
\begin{small}
\begin{equation}\label{know_enc1}
\begin{aligned}
&ww_{i,j} = \sigma\left((f_{1}\left([\bm{w}_{i}, \bm{w}_{j}]\right)+g_{i,j})/\tau\right), \\
&wo_{i,c} = \sigma\left((f_{2}\left([\bm{w}_{i}, \bm{o}_{c}]\right)+g_{i,c})/\tau\right),
\end{aligned}
\end{equation}
\end{small}

\noindent where $\{g_{i,j},g_{i,c}\}$ are i.i.d. sampled from $\operatorname{Gumbel}(0,1)$ distribution~\cite{jang2016categorical}. $\tau$ is a temperature parameter that controls the degree of approximation.
\subsubsection{Knowledge Decoder.}
Knowledge Decoder $P_{\theta}(Y|Z,X)$ applies the knowledge $Z$ acquired by Knowledge Encoder to reason expressions $Y$, which operates the \emph{knowledge-expression} process in Figure~\ref{leap}. Here, we aim to design a general knowledge application mechanism that benefits different solvers (e.g., Seq2Seq, GTS), rather than propose a special solver architecture. Specifically, we contribute a semantics-enhanced module and a reasoning-enhanced module in Knowledge Decoder, which apply knowledge to improve problem understanding and symbol reasoning, respectively. In the following, we first unify the architecture of most existing solvers into ``Backbone Solver''. Then, we explain the details of our proposed modules.

\textbf{Backbone Solver.}
Given problem $P$, a backbone solver first reads the problem sentence $X_P=\{w_1,...,w_n\}$, and then generates word representations $\mathbf{H}=\{\bm{h}_1,...,\bm{h}_n\}$ and an initial reasoning state $\bm{s}_1$ by:
\begin{small}
\begin{equation}\label{se_1}
(\mathbf{H}, \bm{s}_{1})=\operatorname{Sol-Enc}(\{\bm{w}_{i},i=1,...,n\}).
\end{equation}
\end{small}

$\operatorname{Sol-Enc}$ conducts problem understanding and captures many existing models in different solvers. It can be formalized ranging from RNN (e.g., GTS, TSN-MD), BERT (e.g., MWP-BERT), to specific MWP encoder (e.g., Graph2Tree). Then, the backbone solver generates the expression $Y_P=\{y_1,...,y_m\}$ for $P$ step by step. Specifically, at step $t$ ($t=1,...,m$), it reasons symbol $y_t$ by:
\begin{small}
\begin{equation}\label{sd_1}
  P_\theta(y_{t} \mid y_{1},..., y_{t-1}, P)=\operatorname{Sol-Dec}(\bm{s}_t,\bm{e}(y_t),\emph{options}).
\end{equation}
\end{small}

In $\operatorname{Sol-Dec}$, $\bm{s}_t$ is the reasoning state at step $t$ ($\bm{s}_1$ at step 1 comes from Eq.~(\ref{se_1})). $\bm{e}(y_t)$ is the embedding of symbol $y_t$, which equals $\bm{o}_c$ if $y_t$ is operator $o_c$, or $\bm{h}_i$ if $y_t$ is a number $w_i\in N_P$. $\emph{options}$ are some optional terms. Specifically, the meanings of $\operatorname{Sol-Dec}$, $\bm{s}_t$, and $\emph{options}$ vary with different solvers. For example, in Seq2Seq, $\operatorname{Sol-Dec}$ represents the output gate of LSTM, $\bm{s}_t$ is the hidden state, and $\emph{options}=\emptyset$.

After generating symbol $y_t$, the backbone solver derives the next reasoning state $\bm{s}_{t+1}$ with another solver-specific network $f_3$ (e.g., $f_3$ represents the forget gate in Seq2Seq):
\begin{small}
\begin{equation}\label{sd_1_0}
  \bm{s}_{t+1}=f_{3}(\bm{s}_{t}, \bm{e}(y_{t}),\emph{options}).
\end{equation}
\end{small}

Readers can refer to the original papers for more details of different backbone solvers summarized in Section~\ref{related_work}.

\textbf{Semantics-enhanced Module.} LeAp improves problem comprehension with a semantics-enhanced module for application of word-word knowledge $ww_{i,j}$ $\in$ $Z$ (e.g., understand ``how many fruits'' in Figure~\ref{intro} via the knowledge ``apples'' belongs to ``fruits''), which is independent of specific backbone solvers. Specifically, for problem $P$, (1) we construct a graph $SE_P$ as shown in Figure~\ref{leap} by taking the words in $P$ as vertices and their knowledge $\{ww_{i,j}|i,j=1,...,n\}$ from $Z$ as edges. Thus, $SE_P$ can be seen as a subgraph of our overall explicit knowledge $Z$; (2) we get word representations $\mathbf{H}$ by the original backbone according to Eq.~\eqref{se_1}; (3) we utilize a GCN to pass the message on $SE_P$ to obtain $\mathbf{H}^{\prime}$ that fuses the relational knowledge:
\begin{small}
\begin{equation}\label{sd_1_1}
\mathbf{H}^{\prime}=\mathbf{A_E}\cdot \operatorname{ReLU}(\mathbf{A_E}\cdot \mathbf{H}\cdot \mathbf{W}_1+b_1)\cdot \mathbf{W}_2+b_2,
\end{equation}
\end{small}

\noindent where $\mathbf{A_E}\in \mathbb{R}^{n \times n}$ is the adjacency matrix of $SE_P$, $\mathbf{W}_1,\mathbf{W}_2,b_1,b_2$ are trainable parameters.

\textbf{Reasoning-enhanced Module.} Now, we introduce how LeAp improves symbol reasoning with our proposed reasoning-enhanced module that applies both word-word knowledge and word-operator knowledge $ww_{i,j},wo_{i,c}\in Z$. Intuitively, word ``times'' in a problem sentence can encourage a solver to correctly reason symbol ``$\times$'' through the relational knowledge $wo_{i,c}$ between ``times'' and ``$\times$'', whose location in the expression can be refined based on how focused the solver is on ``times''. Thus, to guide the reasoning at step $t$, we first establish temporary relationships $ws^t_i$ between current reasoning state $s_t$ and all words in $P$ by:
\begin{small}
\begin{equation}\label{ws}
\begin{gathered}
ws^t_i=\frac{\exp \left(f_4\left(\bm{s}_t, \bm{h}_i^{\prime}\right)\right)}{\sum_{j} \exp \left(f_4\left(\bm{s}_t, \bm{h}_j^{\prime}\right)\right)}, \\
f_4\left(\bm{s}_t, \bm{h}_i^{\prime}\right)=\bm{v}^{\top} \tanh \left(\mathbf{W}_{3}\cdot\left[\bm{s}_t, \bm{h}_i^{\prime}\right]\right),
\end{gathered}
\end{equation}
\end{small}

\noindent where $\bm{v}$, $\mathbf{W}_{3}$ are learnable parameters.

Then, combining $s_t$ and knowledge $Z$, we construct another graph $RE^t_P$ for problem $P$. As depicted in Figure~\ref{leap}, its vertex set contains reasoning state $\bm{s}_t$, words $\mathbf{H^{\prime}}=\{\bm{h}^{\prime}_1,...,\bm{h}^{\prime}_n\}$ of $P$, and all operators $\mathbf{O}=\{\bm{o}_1,...,\bm{o}_C\}$. The edge set is composed of $ws^t_i$ and knowledge $ww_{i,j}$, $wo_{i,c}$. On $RE^t_P$, we propagate the information from $\bm{s}_t$ to enhance operator representations by:
\begin{small}
\begin{align}
\mathbf{H}^{t} = \mathbf{A_E}\cdot \operatorname{ReLU}(\mathbf{A_E}\cdot [ws^t\cdot\bm{s}_t, \mathbf{H^{\prime}}]\cdot \mathbf{W}_4&+b_4)\cdot \mathbf{W}_5+b_5,\notag\\
\hat{\mathbf{H}}^{t}=\mathbf{H}^{t}+\operatorname{LayerNorm}(\mathbf{H}^{t}&), \notag\\
\mathbf{O}^t = \operatorname{LayerNorm}(\operatorname{ReLU}(\mathbf{A_D}\cdot \hat{\mathbf{H}}^{t}\cdot &\mathbf{W}_6+b_6)),\label{sd_2}\\
\hat{\mathbf{O}}^{t}=\operatorname{ReLU}([\mathbf{O}^t,\mathbf{O}]\cdot \mathbf{W}_7+&b_7),\notag\\
\mathbf{O}^{\prime}=\mathbf{O}+\operatorname{LayerNorm}(\hat{\mathbf{O}}^{t}&),\notag
\end{align}
\end{small}

\noindent where $\mathbf{A_D}\in \mathbb{R}^{C \times n}$ is the adjacency matrix of $RE^t_P$, $\mathbf{W}_*,b_*$ are weight matrices and biases.

When reasoning symbol $y_t$ in Eq.~(\ref{sd_1}), $\bm{e}(y_t)$ is chosen as $\bm{o}^{\prime}_c\in \mathbf{O}^{\prime}$ in Eq.~(\ref{sd_2}) if $y_t$ is operator $o_c$, or $\bm{h}^{\prime}_i\in\mathbf{H}^{\prime}$ in Eq.~(\ref{sd_1_1}) if $y_t$ is number $w_i$. Finally, the symbol representation $\bm{e}(y_t)$ enhanced with knowledge $Z$ is used to generate the next reasoning state $\bm{s}_{t+1}$ by Eq.~(\ref{sd_1_0}), and $RE^t_P$ evolves to $RE^{t+1}_P$.
\subsubsection{Prior of Knowledge.}\label{prior_know}
The prior $P(Z)$ controls the known information for LeAp. Since knowledge $ww_{i,j},wo_{i,c}\in Z$ is binary, it is natural to take the following Bernoulli distribution as the prior of $Z$ and set $\delta_1=0.1$ for sparsity.
\begin{small}
\begin{equation}\label{prior}
 ww_{i,j}\sim \operatorname{Bernoulli}(\delta_1),\ wo_{i,c}\sim \operatorname{Bernoulli}(\delta_1).
\end{equation}
\end{small}

We design the prior $P(Z)$ to simulate a real learner. In practice, a learner may obtain different reasoning performances for MWP with different knowledge backgrounds. For example, a junior high school student has better reasoning ability than a primary school student. Thus, we can set a higher $\delta_1=0.5$ for knowledge that a learner has mastered, which we can simulate by introducing part of (e.g., $\alpha=20\%$) edges of an external knowledge base. With such prior, LeAp can also be guided to learn similar knowledge, thus alleviating the problem of capturing false correlations.

In summary, our LeAp framework has the following advantages. First, LeAp is general to take different MWP solvers as the backbone, enabling them not only to learn reasonable knowledge but also to gain better reasoning ability. Second, LeAp adopts a \emph{problem-knowledge-expression} architecture, where knowledge $Z$ explicitly forms a knowledge graph. Therefore, both its learning results and applying methods are more interpretable compared with previous methods that follow the \emph{problem-expression} paradigm. Third, in the prior of knowledge $Z$, we can set different $\alpha$ to investigate problem solving effects of learners with variable background, which is further visualized in Section~\ref{klp}.
\section{Theoretical Analyses}\label{theo_analysis}
In brief, our LeAp learns mathematical knowledge from problems and applies it to solve MWP, naturally constructing an explicit knowledge graph (KG) as shown in Figure~\ref{leap}. Comparatively, a straightforward way to learn a KG is the link prediction task (LP)~\cite{chen2021topology,cai2020multi}, which predicts the edge between each pair of vertices directly. In this section, we investigate deeper into how LeAp is superior in its autonomous learning mechanism compared with LP. Here, we unify some important notations without loss of generality. Specifically, we use $z_{i,j}\in Z$ to represent the knowledge between vertices $i$ and $j$ in the KG, including word-word relation $ww_{i,j}$ and word-operator relation $wo_{i,c}$. $X$ represents the embeddings of vertices, including words $\{\bm{w}_i\}$ and operators $\{\bm{o}_c\}$.

Before detailed derivation, we first assume that the knowledge $Z$ is helpful for solving MWP, because we cannot expect to gain knowledge irrelevant to mathematical reasoning (e.g., chemistry knowledge) from solving mathematical problems. Thus, we give the following definition:

\textbf{Definition 1.} \emph{Effective knowledge} $Z$: $\forall z_{i,j}\in Z,$
\begin{small}
\begin{equation}\label{eff_know}
[P_{\theta}(Y|z_{i,j}=1,X)-P_{\theta}(Y|z_{i,j}=0,X)]\cdot (2r_{i,j}-1)>0.
\end{equation}
\end{small}
\indent $r_{i,j}$ is the ground-truth label of $z_{i,j}$, equalling $1$ if there exists true knowledge (i.e., edge) between $i$ and $j$, and $0$ otherwise. $P_{\theta}$ is the Knowledge Decoder of LeAp. In Definition 1, if there exists knowledge between $i$ and $j$ (i.e., $r_{i,j}=1$), Knowledge Decoder can better reason the expressions $Y$ by applying $z_{i,j}=1$ than $z_{i,j}=0$. On the contrary, it will introduce redundancy if we treat the false knowledge (i.e., $r_{i,j}=0$) as true (i.e., $z_{i,j}=1$), and thus $P_{\theta}(Y|z_{i,j}=1,X)<P_{\theta}(Y|z_{i,j}=0,X)$ at this time.

Based on the ``Effective knowledge'' assumption, we analyze the difference between our LeAp and LP. Recall that LP learns knowledge with a model $P_{\varphi}(z_{i,j}|X)$, which calculates the posterior probability of $Z$ given $X$ from a Bayesian perspective. Comparatively, LeAp learns knowledge from reasoning expressions $Y$ based on $X$. Thus, we investigate
\begin{itemize}
  \item (1) \textbf{Whether the optimization objective of LeAp in Eq.~(\ref{elbo}) optimizes the posterior $\bm{P_{\varphi}(z_{i,j}|X,Y)}$}
  \item (2) \textbf{If (1) holds, whether $\bm{P_{\varphi}(z_{i,j}|X,Y)}$ is larger (i.e., more accurate) than $\bm{P_{\varphi}(z_{i,j}|X)}$}
\end{itemize}

Here, we consider a simplified setting to train LeAp with parameters being initialized with $\varphi^{LP}, X^{LP}$ from a trained LP model. Under such setting, we have two theorems that answer the two questions above, respectively.

\textbf{Theorem 1.} Assume $P_{\varphi}(z_{i,j}=r_{i,j}|X)>\delta(X)$ holds in a neighborhood $U$ of $(\varphi^{LP}, X^{LP})$ and knowledge $Z$ is effective. Then, for each $z_{i,j} \in Z$, maximizing the objective of MWP solving in Eq.~(\ref{elbo}), i.e., $L_1=\mathrm{E}_{P_{\varphi}(z_{i,j} |X)}[\log P_{\theta}(Y |z_{i,j},X)]$, is equivalent to maximizing
\begin{small}
\begin{align}
L_{3}&=r_{i, j} \cdot P_{\theta}(Y | z_{i, j}=1, X) \cdot P_{\varphi}(z_{i, j}=1 | X)+\notag\\
&(1-r_{i, j}) \cdot P_{\theta}(Y | z_{i, j}=0, X) \cdot P_{\varphi}(z_{i, j}=0 | X)\label{l3}
\end{align}
\end{small}
in $U$, where $\delta(X)\triangleq\max\ \{\frac{1}{1+\frac{\beta(1-r_{i,j},r_{i,j},c)}{\beta(r_{i,j},1-r_{i,j},c)}}\}|_{c=\theta,\bm{x}_i\in X}$, and $\beta(a,b,c)\triangleq P_{\theta}(Y|z_{i, j}=a, X) \cdot \|\frac{\partial P_{\theta}(Y|z_{i, j}=b, X)}{\partial c}\|$.

\emph{Proof}. The basic idea is to verify that the inner product of the derivatives of $L_1$ and $L_3$ is positive for each parameter in LeAp. Thus, under a first order Taylor approximation, the gradient direction of $L_1$ implicitly optimizes $L_3$. On this basis, the entire proof consists of three parts of verification for Knowledge Encoder $\varphi$, Knowledge Decoder $\theta$, and vertex embeddings $X$ respectively. Here $\langle, \rangle$ represents the inner product of two vectors.

1) For $\varphi$ in Knowledge Encoder, $\langle\frac{\partial L_{1}}{\partial \varphi}, \frac{\partial L_{3}}{\partial \varphi}\rangle=$
\begin{small}
\begin{equation}\label{varphi_pro}
(2r_{i, j}-1) \cdot \ln \frac{P_{\theta}(Y | z_{i, j}=1, X)}{P_{\theta}(Y |z_{i, j}=0, X)}\cdot \|\frac{\partial P_{\varphi}(z_{i, j}=1| X)}{\partial \varphi}\|^{2}.
\end{equation}
\end{small}

According to the definition of ``Effective knowledge'' in Eq.~\eqref{eff_know}, $\langle\frac{\partial L_{1}}{\partial \varphi}, \frac{\partial L_{3}}{\partial \varphi}\rangle \geq 0$ always holds.

2) For $\theta$ in Knowledge Decoder, when $r_{i,j}=1$, we can derive that $\langle\frac{\partial L_{1}}{\partial \theta}, \frac{\partial L_{3}}{\partial \theta}\rangle=$
\begin{small}
\begin{align}
& \|\frac{\partial P_{\theta}(Y|z_{i, j}=1, X)}{\partial \theta}\|^{2}\cdot\frac{P_{\varphi}(z_{i, j}=1| X)^2}{P_{\theta}(Y|z_{i, j}=1, X)}+ \notag\\
& \langle\frac{\partial P_{\theta}(Y|z_{i, j}=0, X)}{\partial \theta},\frac{\partial P_{\theta}(Y|z_{i, j}=1, X)}{\partial \theta}\rangle \cdot\notag\\
& \frac{P_{\varphi}(z_{i, j}=0| X)\cdot P_{\varphi}(z_{i, j}=1| X)}{P_{\theta}(Y|z_{i, j}=0, X)}>\notag\\
& \|\frac{\partial P_{\theta}(Y|z_{i, j}=1, X)}{\partial \theta}\| \cdot P_{\varphi}(z_{i, j}=1| X)\cdot \beta(1,0,\theta)\cdot\notag\\
& \frac{P_{\varphi}(z_{i, j}=1| X)\cdot \frac{\beta(0,1,\theta)}{\beta(1,0,\theta)} -P_{\varphi}(z_{i, j}=0| X)}{P_{\theta}(Y|z_{i, j}=1, X)\cdot P_{\theta}(Y|z_{i, j}=0, X)}.\label{theta_pro}
\end{align}
\end{small}
Given the condition $P_{\varphi}(z_{i,j}=r_{i,j}|X)>\delta(X)$, it is easy to verify that $P_{\varphi}(z_{i, j}=1| X)\cdot \frac{\beta(0,1,\theta)}{\beta(1,0,\theta)} >P_{\varphi}(z_{i, j}=0| X)$, and thus $\langle\frac{\partial L_{1}}{\partial \theta}, \frac{\partial L_{3}}{\partial \theta}\rangle>0$ in $U$. The proof for $r_{i,j}=0$ is similar.

3) For $\bm{x}_i \in X$ , we also report the result of $r_{i,j}=1$. Since LeAp is initialized with a trained LP model, $(\varphi^{LP},X^{LP})$ can be seen as a locally optimal solution of $P_{\varphi}(z_{i, j}=1| X)$, and thus $\frac{\partial P_{\varphi}(z_{i, j}=1| X)}{\partial \bm{x}_i}\approx 0$ holds in $U$. Based on it, we can derive that $\langle\frac{\partial L_{1}}{\partial \bm{x}_i}, \frac{\partial L_{3}}{\partial \bm{x}_i}\rangle\approx$
\begin{table*}
\small
\centering
\begin{tabular}{c|ccccccccccc}
\hline
& \multicolumn{3}{c} {Math23K} && \multicolumn{3}{c} {MAWPS} && \multicolumn{3}{c} {SVAMP}\\
\cline { 2 - 4 }\cline { 6 - 8 }\cline { 10 - 12 } & ORI & LeAp & LeAp-EK && ORI& LeAp & LeAp-EK && ORI& LeAp & LeAp-EK\\
\hline
Seq2Seq & $0.640$ & $\mathbf{0.660}^{**}$ & $0.652^{**}$ && $0.797$ & $0.803$ & $\mathbf{0.807}^{*}$ && $0.200$ & $\mathbf{0.236}^{***}$ & $0.220^{***}$ \\
Graph2Tree & $0.774$ & $0.779^{*}$ & $\mathbf{0.782}^{**}$ && $0.837$ & $\mathbf{0.852}^{**}$ & $0.849^{**}$ && $0.319$ & $\mathbf{0.341}^{***}$ & $0.325^{*}$\\
HMS & $0.761$ & $\mathbf{0.769}$ & $0.765$ && $0.803$ & $\mathbf{0.812}^*$ & $0.805$ && $0.179$ & $\mathbf{0.196}^{**}$ & $0.191^{**}$ \\
GTS & $0.756$ & $\mathbf{0.772}^{**}$ & $0.767^{**}$ && $0.826$ & $\mathbf{0.834}^{**}$ & $0.830^{*}$ && $0.277$ & $\mathbf{0.285}$ & $0.279$ \\
TSN-MD & $0.774$ & $\mathbf{0.786}^{**}$ & $0.778^{*}$ && $0.844$ & $\mathbf{0.853}^*$ & $0.848$ && $0.290^{\dag}$ & $\mathbf{0.302}^{**}$ & $0.294^{*}$\\
Multi-E/D & $0.784$ & $0.791^{*}$ & $\mathbf{0.793}^{**}$&& $/$ & $/$ & $/$ && $/$ & $/$ & $/$\\
\hline
\end{tabular}
\caption{Answer accuracy ($***: p\leq 0.001,**: p\leq 0.01,*: p\leq 0.05$). $\dag$: implemented by MTPToolkit~\cite{lan2022mwptoolkit}.}\label{result1}
\end{table*}
\begin{table*}
\small
\centering
\begin{tabular}{c|cccccc}
\hline
LeAp& \multicolumn{2}{c} {Math23K} & \multicolumn{2}{c} {MAWPS} & \multicolumn{2}{c} {SVAMP}\\
\cline { 2-3}\cline { 4-5}\cline { 6-7 } (backbone)& w/o SE & w/o RE & w/o SE & w/o RE & w/o SE & w/o RE\\
\hline
Seq2Seq & $0.645$ & $0.648$ & $0.798$ & $0.802$ & $0.210$ & $0.228$\\
Graph2Tree & $0.778$ & $0.776$ & $0.846$ & $0.845$ & $0.335$ & $0.328$\\
HMS & $0.766$ & $0.762$ & $0.809$ & $0.801$ & $0.193$ & $0.189$\\
GTS & $0.770$ & $0.761$ & $0.830$ & $0.830$ & $0.265$ & $0.284$\\
TSN-MD & $0.782$ & $0.783$ & $0.847$ & $0.852$ & $0.293$ & $0.300$\\
Multi-E/D & $0.790$ & $0.788$ & $/$ & $/$ & $/$ & $/$\\
\hline
\end{tabular}
\caption{Ablation study on reducing semantics-enhanced module (``w/o SE'') or reasoning-enhanced module (``w/o RE''). The performance of LeAp is referred in Table~\ref{result1}.}\label{result2}
\end{table*}
\begin{small}
\begin{align}
& \|\frac{\partial P_{\theta}(Y|z_{i, j}=1, X)}{\partial \bm{x}_i}\|^2\cdot \frac{P_{\varphi}(z_{i, j}=1| X)^2}{P_{\theta}(Y|z_{i, j}=1, X)} +\notag\\
&\langle\frac{\partial P_{\theta}(Y|z_{i, j}=1, X)}{\partial \bm{x}_i}, \frac{\partial P_{\theta}(Y|z_{i, j}=0, X)}{\partial \bm{x}_i}\rangle \cdot\notag\\
& \frac{P_{\varphi}(z_{i, j}=0| X)\cdot P_{\varphi}(z_{i, j}=1| X)}{P_{\theta}(Y|z_{i, j}=0, X)}>\notag\\
& \|\frac{\partial P_{\theta}(Y|z_{i, j}=1, X)}{\partial \bm{x}_i}\| \cdot P_{\varphi}(z_{i, j}=1| X)\cdot \beta(1,0,\bm{x}_i)\cdot\notag\\
& \frac{P_{\varphi}(z_{i, j}=1| X)\cdot \frac{\beta(0,1,\bm{x}_i)}{\beta(1,0,\bm{x}_i)} -P_{\varphi}(z_{i, j}=0| X)}{P_{\theta}(Y|z_{i, j}=1, X)\cdot P_{\theta}(Y|z_{i, j}=0, X)}.\label{x_pro}
\end{align}
\end{small}

Similar to Eq.~(\ref{theta_pro}), it is easy to verify that Eq.~\eqref{x_pro} is larger than $0$, and thus $\langle\frac{\partial L_{1}}{\partial \bm{x}_i}, \frac{\partial L_{3}}{\partial \bm{x}_i}\rangle>0$ in $U$. $\Box$

In Theorem 1, we further notice that $L_3$ can be rewritten as $P(Y|z_{i,j}=r_{i,j},X)\cdot P(z_{i,j}=r_{i,j}|X)$ if $P_{\varphi}(z_{i,j}|X)$ and $P_{\theta}(Y |z_{i, j}, X)$ reconstruct the true distribution $P(z_{i,j}|X)$ and $P(Y |z_{i, j}, X)$ behind dataset $D$. Thus, $L_3$ is equivalent to the following posterior, answering our question (1):
\begin{small}
\begin{equation}\label{post}
\begin{aligned}
&P(z_{i,j}=r_{i,j}|X,Y)\\
&=\frac{P(Y|z_{i,j}=r_{i,j},X)\cdot P(z_{i,j}=r_{i,j}|X)\cdot P(X)}{P(X,Y)}.
\end{aligned}
\end{equation}
\end{small}

\textbf{Theorem 2.} Under the assumption of ``Effective knowledge'', the following inequality holds:
\begin{small}
\begin{equation}\label{post1}
\frac{P(Y|z_{i,j}=r_{i,j},X)\cdot P(X)}{P(X,Y)}>1.
\end{equation}
\end{small}

\emph{Proof}. $P(X,Y)$ can be rewritten as
\begin{small}
\begin{equation}\label{x_y}
\begin{aligned}
&P(X)\cdot [P(Y|z_{i,j}=r_{i,j},X)\cdot P(z_{i,j}=r_{i,j}|X)+\\
&P(Y|z_{i,j}=1-r_{i,j},X)\cdot P(z_{i,j}=1-r_{i,j}|X)].
\end{aligned}
\end{equation}
\end{small}

According to the ``Effective knowledge'' assumption, $P(Y|z_{i,j}=1-r_{i,j},X)<P(Y|z_{i,j}=r_{i,j},X)$. Therefore, Eq.~\eqref{x_y} is less than
\begin{small}
\begin{equation}\label{x_y_1}
\begin{aligned}
&P(X)\cdot [P(Y|z_{i,j}=r_{i,j},X)\cdot P(z_{i,j}=r_{i,j}|X)+\\
&P(Y|z_{i,j}=r_{i,j},X)\cdot P(z_{i,j}=1-r_{i,j}|X)]\\
&=P(X)\cdot P(Y|z_{i,j}=r_{i,j},X),
\end{aligned}
\end{equation}
\end{small}
i.e., inequality.~\eqref{post1} holds. $\Box$

Based on Theorem 2 and Eq.~\eqref{post}, $P(z_{i,j}=r_{i,j}|X,Y)>P(z_{i,j}=r_{i,j}|X)$, supporting our question (2). In summary, we conclude that: According to Theorem 1, in LeAp, the autonomous mechanism of learning knowledge from solving MWP is to optimize the posterior $P(z_{i,j}=r_{i,j}|X,Y)$ by $L_1$ in Eq.~\eqref{elbo}. Such a mechanism is more accurate than a straightforward link prediction model $P(z_{i,j}=r_{i,j}|X)$ since it achieves a higher probability according to Theorem 2. Notably, LeAp's superiority lies in the mechanism to calculate the posterior probability based on the information (i.e., $Y$) provided by solving MWP, instead of obtaining better parameters $\varphi, X$.
\section{Experiments}
\subsection{Experimental Setup}
\textbf{Datasets.} We use three datasets in experiments: \textbf{Math23K}, \textbf{MAWP}, and \textbf{SVAMP}. Specifically, \textbf{Math23K}~\cite{wang2017deep} is a Chinese dataset that contains $23,162$ problems that have only one variable. We use its the published dataset partition for experiments. \textbf{MAWPS}~\cite{roy2017unit} is an English dataset. We select $2,373$ problems with only one unknown variable and conduct 5-fold cross-validation. \textbf{SVAMP}~\cite{patel2021nlp} is a test dataset that contains $1,000$ more difficult problems than MAWPS. Following Patel et al.~\shortcite{patel2021nlp}, models are trained on the combination of MAWPS and another ASDiv-A dataset and tested on the problems in SVAMP.

\noindent \textbf{Baselines.} We take the following SOTA models as the baselines, i.e., the basic Seq2Seq, semantics-focused methods (Graph2Tree, HMS), reasoning-focused methods (GTS, TSN-MD), and ensemble-based method (Multi-E/D). Our LeAp is a general framework, and thus we take all of them as the backbones to evaluate its effectiveness and generality.
\begin{itemize}
  \item \textbf{Seq2Seq}~\cite{luong2015effective} adopts a BiLSTM encoder and a LSTM decoder with attention to translate a problem sentence into an expression.
  \item \textbf{Graph2Tree}~\cite{zhang2020graph} builds and encodes two quantity-related graphs to enrich problem understanding with quantity information.
  \item \textbf{HMS}~\cite{lin2021hms} captures problem semantics following a word-clause-problem hierarchy.
  \item \textbf{GTS}~\cite{xie2019goal} proposes a goal-driven decomposition mechanism to reason the expression tree.
  \item \textbf{TSN-MD}~\cite{zhang2020teacher} generates diverse candidate expressions with a multiple-decoder network.
  \item \textbf{Multi-E/D}~\cite{shen2020solving} is an ensemble of sequence-based encoder/decoder with graph-based encoder/decoder to obtain better semantics and reasoning.
\end{itemize}
\textbf{Implementation Details.} For Knowledge Encoder $P_{\varphi}$, the dimension of embeddings $d$ is $128$. The temperature parameter $\tau$ in Eq.~\eqref{know_enc1} decreases from $0.5$ to $0.1$ during training. For Knowledge Decoder $P_{\theta}$, all backbone solvers are optimized with their original parameter settings. Other parameters are initialized randomly and trained with Adam~\cite{kingma2014adam} with dropout probability $0.5$. Words with less than 5 occurrences are converted to a special token ``UNK''. All experiments are run on a Linux server with four 2.30GHz Intel Xeon Gold 5218 CPUs and a Tesla V100 GPU. Our codes are available at \emph{https://github.com/bigdata-ustc/LeAp}.
\subsection{Performance on Answer Reasoning}
We verify the effectiveness of LeAp in improving the reasoning ability of backbones for MWP, using answer accuracy as the metric since there may be many correct expressions for the same problem. The predicted expression is considered correct if its calculated value equals the true answer.

In Table~\ref{result1}, we first report the performances of all backbones in their original version (``ORI'') and with our LeAp framework (``LeAp''). From Table~\ref{result1}, we observe that LeAp improves the answer accuracy of all backbones, and by applying paired t-test, the improvements are statistically significant with $p\leq 0.001\thicksim 0.05$. It demonstrates that LeAp can enhance the mathematical reasoning ability of MWP solvers by enabling them to autonomously learn explicit knowledge from problem solving and apply it to generate answers.

Second, to assess the effect of knowledge application, we design a variation of LeAp, named ``LeAp-EK'', by directly applying knowledge $Z$ from external knowledge bases in Knowledge Decoder. We select HowNet~\cite{dong2016hownet} as the knowledge base for Math23K and ConceptNet~\cite{speer2017conceptnet} for MAWPS and SVAMP. As we can see, the performances of ``LeAp-EK'' are also significantly better than ``ORI''. It shows the applicability of Knowledge Decoder that applies knowledge for different solvers and further verifies the reasonability of our ``Effective knowledge'' assumption in Section~\ref{theo_analysis}.

Third, ``LeAp'' outperforms ``LeAp-EK'' in almost all cases. That is probably because LeAp learns additional word-operator knowledge and our theoretical results guarantee that effective knowledge for MWP can be acquired. Thus, it reflects the importance and benefits of LeAp's learning mechanism to gain knowledge autonomously.

\textbf{Ablation Study.} Here, we highlight the advantages of our semantics-enhanced module in Eq.~(\ref{sd_1_1}) and reasoning-enhanced module in Eq.~(\ref{sd_2}) for knowledge application. Combing Table~\ref{result1} and Table~\ref{result2}, the accuracy of reducing semantics-enhanced (``w/o SE'') or reasoning-enhanced (``w/o RE'') degrades for all models, proving that they are necessary for LeAp to solve MWP while also having superior generality. Besides, the performances of LeAp ``w/o SE'' are higher than ``w/o RE'' for semantics-focused methods (e.g., Graph2Tree). It indicates that the learned knowledge may be more useful to reason symbols for methods that have captured strong semantics. The opposite results appear for reasoning-focused solvers (e.g., TSN-MD). Thus, these two modules are suitable for different types of backbones.
\begin{table}
\small
\centering
\begin{tabular}{|c|c|c|}
\hline LeAp & \multirow{2}{*}{word-word pairs} & \multirow{2}{*}{word-operator pairs}\\
       (backbone) & & \\
\hline  \multirow{3}{*}{Seq2Seq}  & house-home & total-``$+$''\\
                               & egg-food & selling-``$-$''\\
                               & add-all & pieces-``$\times$''\\
\hline  \multirow{3}{*}{Graph2Tree}  & apple-fruit & more-``$+$''\\
                               & sale-buy & times-``$\times$''\\
                               & give-hold & gave-``$-$''\\
\hline  \multirow{3}{*}{HMS}  & person-people & leftover-``$+$''\\
                               & more-total & other-``$+$''\\
                               & more-another & add-``$+$''\\
\hline  \multirow{3}{*}{GTS}  & potato-food & earned-``$+$''\\
                               & apple-fruit & rate-``$\div$''\\
                               & piece-part & added-``$+$''\\
\hline  \multirow{3}{*}{TSN-MD}  & per-each & give-``$-$''\\
                               & store-sale & costs-``$-$''\\
                               & red-blue & borrowed-``$+$''\\
\hline
\end{tabular}
\caption{Top 3 knowledge (word-word/operator pairs) of LeAp with different backbones on MAWPS.}\label{ww_mawps}
\end{table}
\subsection{LeAp Analysis}\label{klp}
\textbf{Knowledge learning.} Now, we assess the quality of explicit knowledge $Z$ learned by LeAp. Our idea is to rank $ww_{i,j},wo_{i,c}\in Z$ after training and expect the reasonable one to be at the top. We visualize the top 3 word-word/operator pairs in Table~\ref{ww_mawps} and clearly see that LeAp gains reasonable knowledge with different backbones (e.g., LeAp (Seq2Seq) learns that ``house'' is related to ``home'' and ``total'' is related to ``$+$'').

Further, we conduct a quantitative evaluation by introducing $\alpha=20\%$ of external knowledge bases (i.e., HowNet and ConceptNet) as the prior for word-word knowledge $ww_{i,j}$ (there is no external knowledge base for $wo_{i,c}$) and calculating $Precision@50$ based on the rest $80\%$, because we focus more on the knowledge accuracy. Here, We introduce three classic baselines. ``Co-occur'' determines the weight between two words by their co-occurrence frequency in all problems. ``TF-IDF'' calculates the cosine similarity between words by their TF-IDF values. ``LP'' is a model with the same architecture as Knowledge Encoder. It is trained on the same $20\%$ knowledge in the prior of LeAp.

\begin{figure}[t]
\centering
\includegraphics[width=\linewidth]{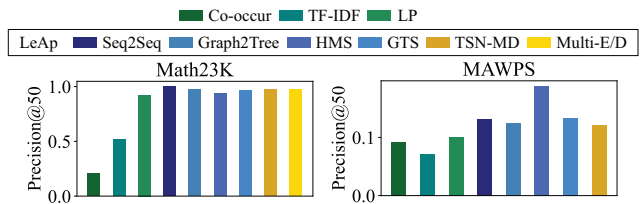}
\caption{$Precision@50$ of word-word relations $ww_{i,j}$.}
\label{know_qual_vis}
\end{figure}
From Figure~\ref{know_qual_vis}, our LeAp performs better than all baselines, no matter taking which backbone solver. It reflects the superiority and robustness of LeAp's autonomous learning mechanism. Especially, LeAp outperforms LP, justifying the rationality of our theoretical analyses in Section~\ref{theo_analysis}.
%
\begin{figure}[t]
\centering
\includegraphics[width=\linewidth]{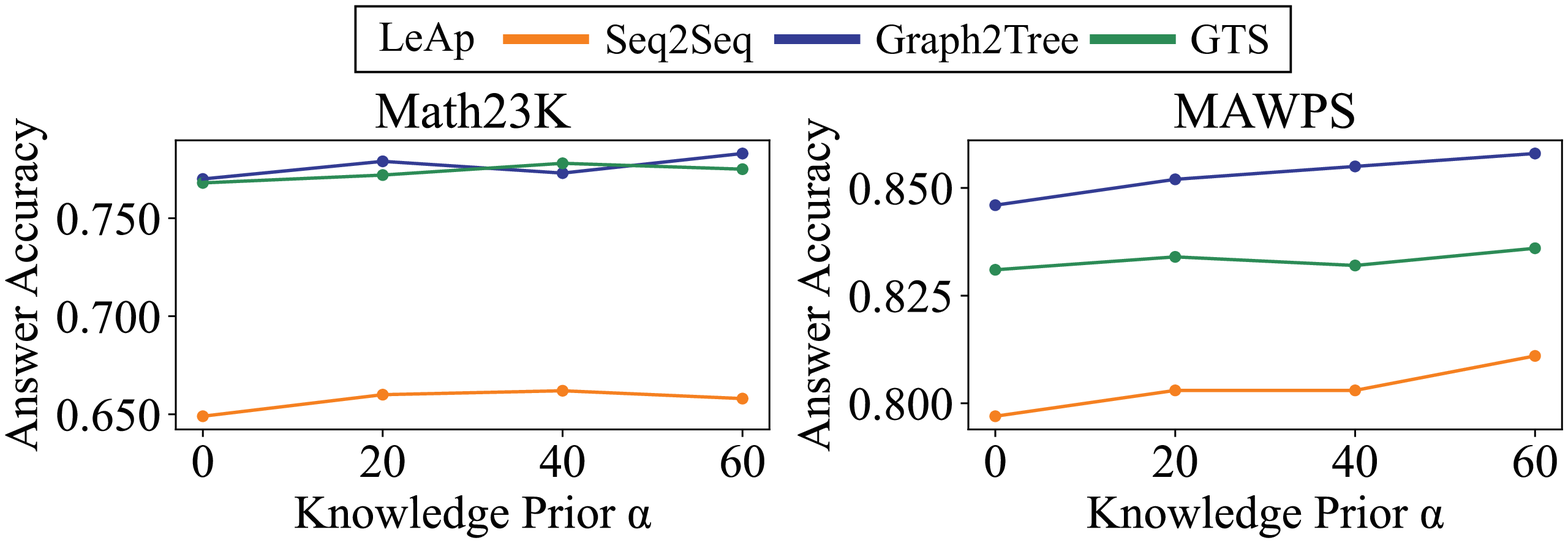}
\caption{Answer accuracy with $\alpha=0\%,20\%,40\%,60\%$.}
\label{prior_know_vis}
\end{figure}
\begin{figure}[t]
\centering
\includegraphics[width=\linewidth]{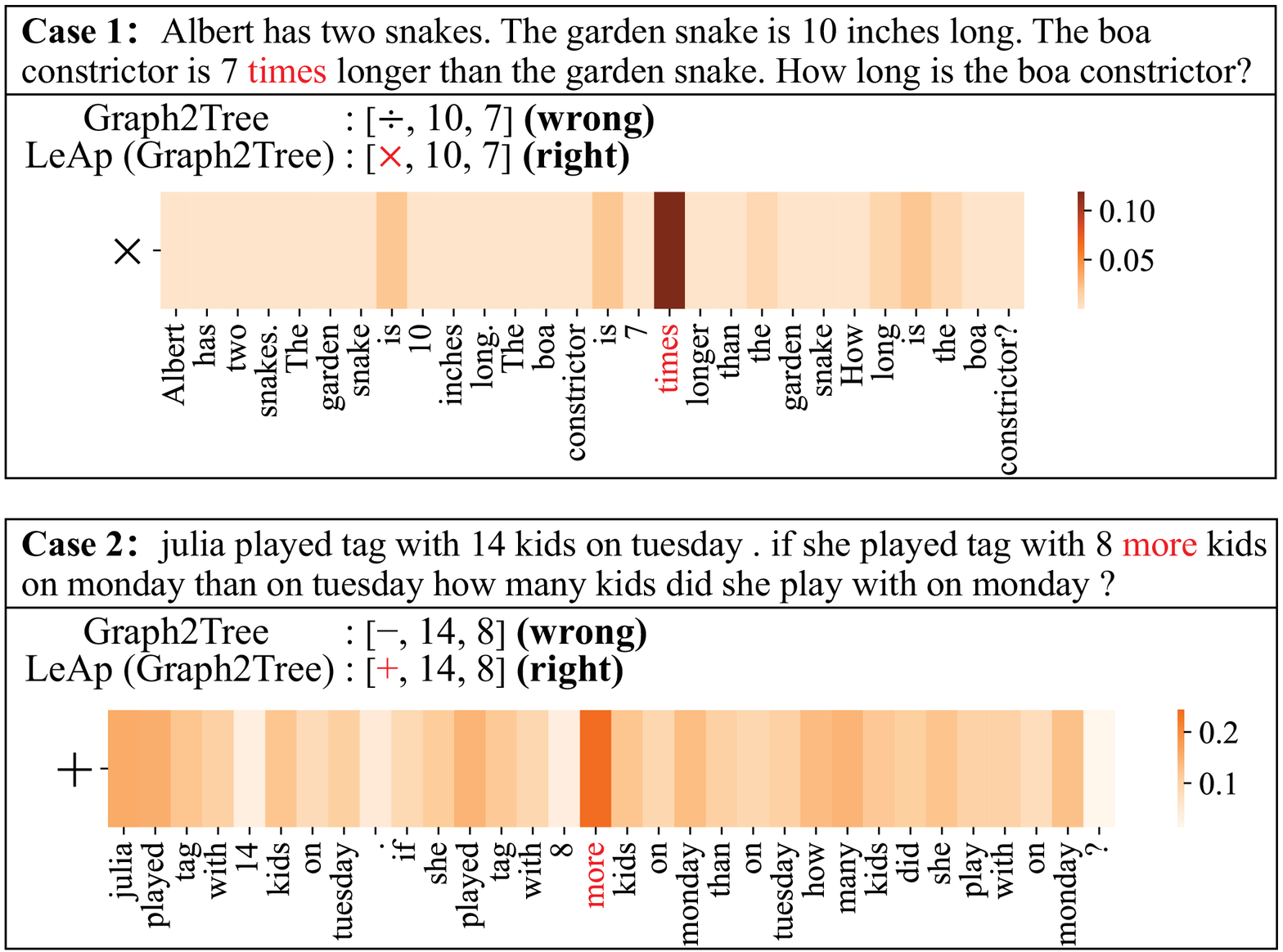}
\caption{Case study.}
\label{case_study}
\end{figure}

\textbf{Knowledge prior.} To simulate learners with different backgrounds, we take $\alpha=0\%,20\%,40\%,60\%$ of external knowledge bases in the prior $P(Z)$ and evaluate the performance of LeAp on MWP solving, with Seq2Seq, Graph2Tree, and GTS as the backbone. From Figure~\ref{prior_know_vis}, with the increase of $\alpha$, the answer accuracy of LeAp shows an increasing trend, just as a human learner with a richer knowledge foundation can better carry out mathematical reasoning. Besides, when $\alpha=0\%$, LeAp still outperforms the original backbones (``ORI'') in Table~\ref{result1}. It demonstrates the flexibility of our LeAp to learn knowledge from scratch.

\textbf{Case Study.} Further, we conduct case study to illustrate the interpretable reasoning process of LeAp (Graph2Tree as the backbone). In Figure~\ref{case_study}, we report the problem sentence, the prefix expressions reasoned by Graph2Tree and LeAp (Graph2Tree), and the word-operator knowledge $wo_{i,c}$ learned by LeAp. For both cases, we can see that the original Graph2Tree makes mistakes at reasoning step $t=1$. Comparatively, by applying the learned knowledge $wo_{i,c}$ between ``times'' and ``$\times$'', ``more'' and ``$+$'', LeAp corrects such errors and reasons ``$\times$'' and ``$+$'' accurately. Notably, the knowledge $wo_{i,c}$ is fixed during the reasoning process. When to use this knowledge is controlled by how focused LeAp is on the words ``times'' and ``more'', measured by $ws_i^t$ in our proposed reasoning-enhanced module (Eq.~(\ref{ws}) in Section 3.2). Therefore, we can conclude that in LeAp, not only does the learned explicit knowledge benefit existing MWP solvers on answer accuracy, but the knowledge application mechanism can also explain how to reason the corresponding answers, showing superior interpretability.

\section{Conclusion and Future Work}
In this paper, we proposed a Learning by Applying (LeAp) framework for explicit knowledge learning and applying in a novel general \emph{problem-knowledge-expression} paradigm that can be added to existing solvers to improve their reasoning ability. We also designed semantics/reasoning-enhanced modules in LeAp to strengthen problem understanding and symbol reasoning by applying knowledge effectively. We theoretically proved the superiority of LeAp's autonomous learning mechanism from a Bayesian perspective. Experiments showed the effectiveness of LeAp in answer reasoning, knowledge learning, and interpretability. In the future, we will extend LeAp to other types of knowledge/problems, and explore its potential to enrich external knowledge bases.

\textbf{Acknowledgement.} This research was partially supported by grants from the National Natural Science Foundation of China (Grants No. 62106244, and 61922073).
\bibliography{bib}

\begin{thebibliography}{61}
\providecommand{\natexlab}[1]{#1}

\bibitem[{Bakman(2007)}]{bakman2007robust}
Bakman, Y. 2007.
\newblock Robust understanding of word problems with extraneous information.
\newblock \emph{arXiv preprint math/0701393}.

\bibitem[{Bansal et~al.(2019)Bansal, Juan, Ravi, and McCallum}]{bansal2019a2n}
Bansal, T.; Juan, D.-C.; Ravi, S.; and McCallum, A. 2019.
\newblock A2N: Attending to neighbors for knowledge graph inference.
\newblock In \emph{Proceedings of the 57th annual meeting of the association
  for computational linguistics}, 4387--4392.

\bibitem[{Bordes et~al.(2013)Bordes, Usunier, Garcia-Duran, Weston, and
  Yakhnenko}]{bordes2013translating}
Bordes, A.; Usunier, N.; Garcia-Duran, A.; Weston, J.; and Yakhnenko, O. 2013.
\newblock Translating embeddings for modeling multi-relational data.
\newblock \emph{Advances in neural information processing systems}, 26.

\bibitem[{Cai and Ji(2020)}]{cai2020multi}
Cai, L.; and Ji, S. 2020.
\newblock A multi-scale approach for graph link prediction.
\newblock In \emph{Proceedings of the AAAI conference on artificial
  intelligence}, volume~34, 3308--3315.

\bibitem[{Cao et~al.(2021)Cao, Hong, Li, and Luo}]{cao2021bottom}
Cao, Y.; Hong, F.; Li, H.; and Luo, P. 2021.
\newblock A Bottom-Up DAG Structure Extraction Model for Math Word Problems.
\newblock In \emph{AAAI}, volume~35, 39--46.

\bibitem[{Chen et~al.(2021)Chen, He, Wu, and Wang}]{chen2021topology}
Chen, J.; He, H.; Wu, F.; and Wang, J. 2021.
\newblock Topology-aware correlations between relations for inductive link
  prediction in knowledge graphs.
\newblock In \emph{Proceedings of the AAAI Conference on Artificial
  Intelligence}, volume~35, 6271--6278.

\bibitem[{Cheng et~al.(2021)Cheng, Yang, Zhang, and Sun}]{cheng2021uniker}
Cheng, K.; Yang, Z.; Zhang, M.; and Sun, Y. 2021.
\newblock UniKER: A Unified Framework for Combining Embedding and Definite Horn
  Rule Reasoning for Knowledge Graph Inference.
\newblock In \emph{Proceedings of the 2021 Conference on Empirical Methods in
  Natural Language Processing}, 9753--9771.

\bibitem[{Dai and Muggleton(2021)}]{dai2021abductive}
Dai, W.-Z.; and Muggleton, S.~H. 2021.
\newblock Abductive knowledge induction from raw data.

\bibitem[{de~Penning et~al.(2011)de~Penning, Garcez, Lamb, and
  Meyer}]{de2011neural}
de~Penning, H. L.~H.; Garcez, A. S.~d.; Lamb, L.~C.; and Meyer, J.-J.~C. 2011.
\newblock A neural-symbolic cognitive agent for online learning and reasoning.
\newblock In \emph{Twenty-Second International Joint Conference on Artificial
  Intelligence}.

\bibitem[{Dong and Dong(2016)}]{dong2016hownet}
Dong, Z.; and Dong, Q. 2016.
\newblock Hownet And The Computation Of Meaning: (With CD-ROM).

\bibitem[{Feigenbaum, Feldman et~al.(1963)}]{feigenbaum1963computers}
Feigenbaum, E.~A.; Feldman, J.; et~al. 1963.
\newblock \emph{Computers and thought}.
\newblock New York McGraw-Hill.

\bibitem[{Fletcher(1985)}]{fletcher1985understanding}
Fletcher, C.~R. 1985.
\newblock Understanding and solving arithmetic word problems: A computer
  simulation.
\newblock \emph{Behavior Research Methods, Instruments, \& Computers}, 17(5):
  565--571.

\bibitem[{Hosseini et~al.(2014)Hosseini, Hajishirzi, Etzioni, and
  Kushman}]{hosseini2014learning}
Hosseini, M.~J.; Hajishirzi, H.; Etzioni, O.; and Kushman, N. 2014.
\newblock Learning to solve arithmetic word problems with verb categorization.
\newblock In \emph{EMNLP}, 523--533. Citeseer.

\bibitem[{Huang et~al.(2021)Huang, Lin, Wang, Liu, Chen, Ma, Su, and
  Tong}]{huang2021disenqnet}
Huang, Z.; Lin, X.; Wang, H.; Liu, Q.; Chen, E.; Ma, J.; Su, Y.; and Tong, W.
  2021.
\newblock Disenqnet: Disentangled representation learning for educational
  questions.
\newblock In \emph{Proceedings of the 27th ACM SIGKDD Conference on Knowledge
  Discovery \& Data Mining}, 696--704.

\bibitem[{Huang et~al.(2020{\natexlab{a}})Huang, Liu, Chen, Wu, Xiao, Chen, Ma,
  and Hu}]{huang2020learning}
Huang, Z.; Liu, Q.; Chen, Y.; Wu, L.; Xiao, K.; Chen, E.; Ma, H.; and Hu, G.
  2020{\natexlab{a}}.
\newblock Learning or forgetting? a dynamic approach for tracking the knowledge
  proficiency of students.
\newblock \emph{ACM Transactions on Information Systems (TOIS)}, 38(2): 1--33.

\bibitem[{Huang et~al.(2020{\natexlab{b}})Huang, Liu, Gao, Wu, Yin, Wang, and
  Chen}]{huang2020neural}
Huang, Z.; Liu, Q.; Gao, W.; Wu, J.; Yin, Y.; Wang, H.; and Chen, E.
  2020{\natexlab{b}}.
\newblock Neural mathematical solver with enhanced formula structure.
\newblock In \emph{Proceedings of the 43rd International ACM SIGIR Conference
  on Research and Development in Information Retrieval}, 1729--1732.

\bibitem[{Jang, Gu, and Poole(2016)}]{jang2016categorical}
Jang, E.; Gu, S.; and Poole, B. 2016.
\newblock Categorical reparameterization with gumbel-softmax.
\newblock \emph{arXiv preprint arXiv:1611.01144}.

\bibitem[{Jie, Li, and Lu(2022)}]{jie2022learning}
Jie, Z.; Li, J.; and Lu, W. 2022.
\newblock Learning to Reason Deductively: Math Word Problem Solving as Complex
  Relation Extraction.
\newblock In \emph{Proceedings of the 60th Annual Meeting of the Association
  for Computational Linguistics}, 5944--5955.

\bibitem[{Kim et~al.(2022)Kim, Hwang, Yoo, and Cheong}]{kim2022improving}
Kim, H.; Hwang, J.; Yoo, T.; and Cheong, Y.-G. 2022.
\newblock Improving a Graph-to-Tree Model for Solving Math Word Problems.
\newblock In \emph{2022 16th International Conference on Ubiquitous Information
  Management and Communication (IMCOM)}, 1--7. IEEE.

\bibitem[{Kingma and Ba(2014)}]{kingma2014adam}
Kingma, D.~P.; and Ba, J. 2014.
\newblock Adam: A method for stochastic optimization.
\newblock \emph{arXiv preprint arXiv:1412.6980}.

\bibitem[{Kingma and Welling(2013)}]{kingma2013auto}
Kingma, D.~P.; and Welling, M. 2013.
\newblock Auto-encoding variational bayes.
\newblock \emph{arXiv preprint arXiv:1312.6114}.

\bibitem[{Kipf et~al.(2018)Kipf, Fetaya, Wang, Welling, and
  Zemel}]{kipf2018neural}
Kipf, T.; Fetaya, E.; Wang, K.-C.; Welling, M.; and Zemel, R. 2018.
\newblock Neural relational inference for interacting systems.
\newblock In \emph{International Conference on Machine Learning}, 2688--2697.
  PMLR.

\bibitem[{Koncel-Kedziorski et~al.(2015)Koncel-Kedziorski, Hajishirzi,
  Sabharwal, Etzioni, and Ang}]{koncel2015parsing}
Koncel-Kedziorski, R.; Hajishirzi, H.; Sabharwal, A.; Etzioni, O.; and Ang,
  S.~D. 2015.
\newblock Parsing algebraic word problems into equations.
\newblock \emph{Transactions of the Association for Computational Linguistics},
  3: 585--597.

\bibitem[{Labhishetty et~al.(2022)Labhishetty, Zhai, Xie, Gong, Sharnagat, and
  Chembolu}]{labhishetty2022differential}
Labhishetty, S.; Zhai, C.; Xie, M.; Gong, L.; Sharnagat, R.; and Chembolu, S.
  2022.
\newblock Differential Query Semantic Analysis: Discovery of Explicit
  Interpretable Knowledge from E-Com Search Logs.
\newblock In \emph{Proceedings of the Fifteenth ACM International Conference on
  Web Search and Data Mining}, 535--543.

\bibitem[{Lan et~al.(2022)Lan, Wang, Zhang, Lan, Dai, Wang, Zhang, and
  Lim}]{lan2022mwptoolkit}
Lan, Y.; Wang, L.; Zhang, Q.; Lan, Y.; Dai, B.~T.; Wang, Y.; Zhang, D.; and
  Lim, E.-P. 2022.
\newblock Mwptoolkit: an open-source framework for deep learning-based math
  word problem solvers.
\newblock In \emph{Proceedings of the AAAI Conference on Artificial
  Intelligence}, volume~36, 13188--13190.

\bibitem[{Lewkowycz et~al.(2022)Lewkowycz, Andreassen, Dohan, Dyer,
  Michalewski, Ramasesh, Slone, Anil, Schlag, Gutman-Solo
  et~al.}]{lewkowycz2022solving}
Lewkowycz, A.; Andreassen, A.; Dohan, D.; Dyer, E.; Michalewski, H.; Ramasesh,
  V.; Slone, A.; Anil, C.; Schlag, I.; Gutman-Solo, T.; et~al. 2022.
\newblock Solving Quantitative Reasoning Problems with Language Models.
\newblock \emph{arXiv preprint arXiv:2206.14858}.

\bibitem[{Liang et~al.(2018)Liang, Hu, Zhang, Lin, and
  Xing}]{liang2018symbolic}
Liang, X.; Hu, Z.; Zhang, H.; Lin, L.; and Xing, E.~P. 2018.
\newblock Symbolic graph reasoning meets convolutions.
\newblock \emph{Advances in Neural Information Processing Systems}, 31.

\bibitem[{Liang et~al.(2021)Liang, Zhang, Shao, and Zhang}]{liang2021mwp}
Liang, Z.; Zhang, J.; Shao, J.; and Zhang, X. 2021.
\newblock Mwp-bert: A strong baseline for math word problems.
\newblock \emph{arXiv preprint arXiv:2107.13435}.

\bibitem[{Lin et~al.(2021)Lin, Huang, Zhao, Chen, Liu, Wang, and
  Wang}]{lin2021hms}
Lin, X.; Huang, Z.; Zhao, H.; Chen, E.; Liu, Q.; Wang, H.; and Wang, S. 2021.
\newblock Hms: A hierarchical solver with dependency-enhanced understanding for
  math word problem.
\newblock In \emph{Proceedings of the AAAI Conference on Artificial
  Intelligence}, volume~35, 4232--4240.

\bibitem[{Liu et~al.(2021)Liu, Du, Ji, Zhai, and Tong}]{liu2021neural}
Liu, L.; Du, B.; Ji, H.; Zhai, C.; and Tong, H. 2021.
\newblock Neural-Answering Logical Queries on Knowledge Graphs.
\newblock In \emph{Proceedings of the 27th ACM SIGKDD Conference on Knowledge
  Discovery \& Data Mining}, 1087--1097.

\bibitem[{Liu et~al.(2019)Liu, Huang, Yin, Chen, Xiong, Su, and
  Hu}]{liu2019ekt}
Liu, Q.; Huang, Z.; Yin, Y.; Chen, E.; Xiong, H.; Su, Y.; and Hu, G. 2019.
\newblock Ekt: Exercise-aware knowledge tracing for student performance
  prediction.
\newblock \emph{IEEE Transactions on Knowledge and Data Engineering}, 33(1):
  100--115.

\bibitem[{Luong, Pham, and Manning(2015)}]{luong2015effective}
Luong, M.-T.; Pham, H.; and Manning, C.~D. 2015.
\newblock Effective Approaches to Attention-based Neural Machine Translation.
\newblock In \emph{Proceedings of the 2015 Conference on Empirical Methods in
  Natural Language Processing}, 1412--1421.

\bibitem[{Mitra and Baral(2016)}]{mitra2016learning}
Mitra, A.; and Baral, C. 2016.
\newblock Learning to use formulas to solve simple arithmetic problems.
\newblock In \emph{Proceedings of the 54th Annual Meeting of the Association
  for Computational Linguistics (Volume 1: Long Papers)}, 2144--2153.

\bibitem[{Mowrer(1960)}]{mowrer1960learning}
Mowrer, O. 1960.
\newblock Learning theory and behavior.

\bibitem[{Muhajirah(2020)}]{muhajirah2020basic}
Muhajirah, M. 2020.
\newblock Basic of Learning Theory: (Behaviorism, Cognitivism, Constructivism,
  and Humanism).
\newblock \emph{International Journal of Asian Education}, 1(1): 37--42.

\bibitem[{Patel, Bhattamishra, and Goyal(2021)}]{patel2021nlp}
Patel, A.; Bhattamishra, S.; and Goyal, N. 2021.
\newblock Are NLP Models really able to Solve Simple Math Word Problems?
\newblock In \emph{Proceedings of the 2021 Conference of the North American
  Chapter of the Association for Computational Linguistics: Human Language
  Technologies}, 2080--2094.

\bibitem[{Pei et~al.(2020)Pei, Wei, Chang, Zhang, and Yang}]{pei2020curvature}
Pei, H.; Wei, B.; Chang, K.; Zhang, C.; and Yang, B. 2020.
\newblock Curvature regularization to prevent distortion in graph embedding.
\newblock \emph{Advances in Neural Information Processing Systems}, 33:
  20779--20790.

\bibitem[{Pei et~al.(2019)Pei, Wei, Chang, Lei, and Yang}]{pei2019geom}
Pei, H.; Wei, B.; Chang, K. C.-C.; Lei, Y.; and Yang, B. 2019.
\newblock Geom-GCN: Geometric Graph Convolutional Networks.
\newblock In \emph{International Conference on Learning Representations}.

\bibitem[{Petroni et~al.(2019)Petroni, Rockt{\"a}schel, Riedel, Lewis, Bakhtin,
  Wu, and Miller}]{petroni2019language}
Petroni, F.; Rockt{\"a}schel, T.; Riedel, S.; Lewis, P.; Bakhtin, A.; Wu, Y.;
  and Miller, A. 2019.
\newblock Language Models as Knowledge Bases?
\newblock In \emph{Proceedings of the 2019 Conference on Empirical Methods in
  Natural Language Processing and the 9th International Joint Conference on
  Natural Language Processing (EMNLP-IJCNLP)}, 2463--2473.

\bibitem[{Peyrard and West(2020)}]{peyrard2020klearn}
Peyrard, M.; and West, R. 2020.
\newblock KLearn: Background Knowledge Inference from Summarization Data.
\newblock In \emph{EMNLP (Findings)}.

\bibitem[{Qu and Tang(2019)}]{qu2019probabilistic}
Qu, M.; and Tang, J. 2019.
\newblock Probabilistic logic neural networks for reasoning.
\newblock \emph{Advances in neural information processing systems}, 32.

\bibitem[{Roy and Roth(2017)}]{roy2017unit}
Roy, S.; and Roth, D. 2017.
\newblock Unit dependency graph and its application to arithmetic word problem
  solving.
\newblock In \emph{AAAI}, volume~31.

\bibitem[{Seo et~al.(2015)Seo, Hajishirzi, Farhadi, Etzioni, and
  Malcolm}]{seo2015solving}
Seo, M.; Hajishirzi, H.; Farhadi, A.; Etzioni, O.; and Malcolm, C. 2015.
\newblock Solving geometry problems: Combining text and diagram interpretation.
\newblock In \emph{EMNLP}, 1466--1476.

\bibitem[{Shehata(2009)}]{shehata2009wordnet}
Shehata, S. 2009.
\newblock A wordnet-based semantic model for enhancing text clustering.
\newblock In \emph{2009 IEEE International Conference on Data Mining
  Workshops}, 477--482. IEEE.

\bibitem[{Shen et~al.(2021)Shen, Yin, Li, Shang, Jiang, Zhang, and
  Liu}]{shen2021generate}
Shen, J.; Yin, Y.; Li, L.; Shang, L.; Jiang, X.; Zhang, M.; and Liu, Q. 2021.
\newblock Generate \& Rank: A Multi-task Framework for Math Word Problems.
\newblock In \emph{Findings of the Association for Computational Linguistics:
  EMNLP 2021}, 2269--2279.

\bibitem[{Shen and Jin(2020)}]{shen2020solving}
Shen, Y.; and Jin, C. 2020.
\newblock Solving math word problems with multi-encoders and multi-decoders.
\newblock In \emph{Proceedings of the 28th International Conference on
  Computational Linguistics}, 2924--2934.

\bibitem[{Shi et~al.(2015)Shi, Wang, Lin, Liu, and Rui}]{shi2015automatically}
Shi, S.; Wang, Y.; Lin, C.-Y.; Liu, X.; and Rui, Y. 2015.
\newblock Automatically solving number word problems by semantic parsing and
  reasoning.
\newblock In \emph{EMNLP}, 1132--1142.

\bibitem[{Speer, Chin, and Havasi(2017)}]{speer2017conceptnet}
Speer, R.; Chin, J.; and Havasi, C. 2017.
\newblock Conceptnet 5.5: An open multilingual graph of general knowledge.
\newblock In \emph{Thirty-first AAAI conference on artificial intelligence}.

\bibitem[{Tsatsou et~al.(2021)Tsatsou, Karageorgos, Dimou, Carb{\'o},
  L{\'o}pez, and Daras}]{tsatsou2021towards}
Tsatsou, D.; Karageorgos, K.; Dimou, A.; Carb{\'o}, J.; L{\'o}pez, J. M.~M.;
  and Daras, P. 2021.
\newblock Towards Unsupervised Knowledge Extraction.
\newblock In \emph{AAAI Spring Symposium: Combining Machine Learning with
  Knowledge Engineering}.

\bibitem[{von Rueden et~al.(2021)von Rueden, Mayer, Beckh, Georgiev,
  Giesselbach, Heese, Kirsch, Walczak, Pfrommer, Pick et~al.}]{von2021informed}
von Rueden, L.; Mayer, S.; Beckh, K.; Georgiev, B.; Giesselbach, S.; Heese, R.;
  Kirsch, B.; Walczak, M.; Pfrommer, J.; Pick, A.; et~al. 2021.
\newblock Informed Machine Learning-A Taxonomy and Survey of Integrating Prior
  Knowledge into Learning Systems.
\newblock \emph{IEEE Transactions on Knowledge \& Data Engineering}, (01):
  1--1.

\bibitem[{Wang et~al.(2019)Wang, Zhang, Zhang, Xu, Gao, Dai, and
  Shen}]{wang2019template}
Wang, L.; Zhang, D.; Zhang, J.; Xu, X.; Gao, L.; Dai, B.~T.; and Shen, H.~T.
  2019.
\newblock Template-based math word problem solvers with recursive neural
  networks.
\newblock In \emph{Proceedings of the AAAI Conference on Artificial
  Intelligence}, volume~33, 7144--7151.

\bibitem[{Wang, Liu, and Shi(2017)}]{wang2017deep}
Wang, Y.; Liu, X.; and Shi, S. 2017.
\newblock Deep neural solver for math word problems.
\newblock In \emph{Proceedings of the 2017 Conference on Empirical Methods in
  Natural Language Processing}, 845--854.

\bibitem[{Wu et~al.(2020)Wu, Zhang, Fu, and Huang}]{wu2020knowledge}
Wu, Q.; Zhang, Q.; Fu, J.; and Huang, X.-J. 2020.
\newblock A knowledge-aware sequence-to-tree network for math word problem
  solving.
\newblock In \emph{EMNLP}, 7137--7146.

\bibitem[{Wu, Zhang, and Wei(2021)}]{wu2021edge}
Wu, Q.; Zhang, Q.; and Wei, Z. 2021.
\newblock An edge-enhanced hierarchical graph-to-tree network for math word
  problem solving.
\newblock In \emph{Findings of the Association for Computational Linguistics:
  EMNLP 2021}, 1473--1482.

\bibitem[{Xia et~al.(2021)Xia, Huang, Xu, Dai, Zhang, Yang, Pei, and
  Bo}]{xia2021knowledge}
Xia, L.; Huang, C.; Xu, Y.; Dai, P.; Zhang, X.; Yang, H.; Pei, J.; and Bo, L.
  2021.
\newblock Knowledge-enhanced hierarchical graph transformer network for
  multi-behavior recommendation.
\newblock In \emph{Proceedings of the AAAI Conference on Artificial
  Intelligence}, volume~35, 4486--4493.

\bibitem[{Xie and Sun(2019)}]{xie2019goal}
Xie, Z.; and Sun, S. 2019.
\newblock A Goal-Driven Tree-Structured Neural Model for Math Word Problems.
\newblock In \emph{IJCAI}, 5299--5305.

\bibitem[{Yu et~al.(2021)Yu, Wen, Zheng, and Xiao}]{yu2021improving}
Yu, W.; Wen, Y.; Zheng, F.; and Xiao, N. 2021.
\newblock Improving Math Word Problems with Pre-trained Knowledge and
  Hierarchical Reasoning.
\newblock In \emph{Proceedings of the 2021 Conference on Empirical Methods in
  Natural Language Processing}, 3384--3394.

\bibitem[{Zhang, Wang et~al.(2020)}]{zhang2020gap}
Zhang, D.; Wang, L.; et~al. 2020.
\newblock The Gap of Semantic Parsing: A Survey on Automatic Math Word Problem
  Solvers.
\newblock \emph{IEEE Transactions on Pattern Analysis and Machine
  Intelligence}, 42(9): 2287--2305.

\bibitem[{Zhang et~al.(2020{\natexlab{a}})Zhang, Lee, Lim, Qin, Wang, Shao, and
  Sun}]{zhang2020teacher}
Zhang, J.; Lee, R. K.-W.; Lim, E.-P.; Qin, W.; Wang, L.; Shao, J.; and Sun, Q.
  2020{\natexlab{a}}.
\newblock Teacher-student networks with multiple decoders for solving math word
  problem.
\newblock In \emph{IJCAI}.

\bibitem[{Zhang et~al.(2020{\natexlab{b}})Zhang, Wang, Lee
  et~al.}]{zhang2020graph}
Zhang, J.; Wang, L.; Lee, R. K.-W.; et~al. 2020{\natexlab{b}}.
\newblock Graph-to-Tree Learning for Solving Math Word Problems.
\newblock In \emph{Association for Computational Linguistics}, 3928--3937.

\bibitem[{Zhou et~al.(2018)Zhou, Young, Huang, Zhao, Xu, and
  Zhu}]{zhou2018commonsense}
Zhou, H.; Young, T.; Huang, M.; Zhao, H.; Xu, J.; and Zhu, X. 2018.
\newblock Commonsense knowledge aware conversation generation with graph
  attention.
\newblock In \emph{IJCAI}, 4623--4629.

\end{thebibliography}

\end{document}